\documentclass[journal]{IEEEtran}
\IEEEoverridecommandlockouts

\usepackage{graphicx}
\usepackage{amsmath}
\usepackage{amssymb}
\usepackage{comment}
\usepackage{xcolor}
\usepackage{bm}
\usepackage{tikz}
\usetikzlibrary{arrows.meta, backgrounds, positioning, fit}
\pgfdeclarelayer{background}
\pgfsetlayers{background,main}
\usepackage{subcaption}
\usepackage[inkscapearea=page]{svg}
\usepackage{multirow}
\usepackage{booktabs}
\usepackage[T1]{fontenc}
\usepackage[ruled,vlined]{algorithm2e}
\usepackage{adjustbox}

\definecolor{colGR}{RGB}{224,168,33}
\definecolor{colIDR}{RGB}{142,78,198}
\definecolor{colDDR}{RGB}{47,158,95}

\definecolor{taskspace}{RGB}{246,247,249}
\definecolor{taskspaceborder}{RGB}{150,155,165}
\definecolor{jointspace}{RGB}{218,220,226}
\definecolor{jointspaceborder}{RGB}{95,100,110}

\definecolor{darkblue}{RGB}{20,55,105}

\usepackage{cite}
\usepackage[colorlinks=true, linkcolor=darkblue, citecolor=darkblue, urlcolor=darkblue]{hyperref}

\newcommand{\R}{\mathbb{R}}

\newcommand{\Q}{\mathcal{Q}}
\newcommand{\F}{\mathcal{F}}
\newcommand{\TF}{\widetilde{\mathcal{F}}}
\newcommand{\U}{\mathcal{U}}
\DeclareMathOperator*{\argmin}{arg\,min}

\title{Direct Dynamic Retargeting for Humanoid Imitation Learning from Videos}
\author{Constant Roux$^{\star, 1}$, Ludovic De Matteïs$^{\star, 1}$, Mathias Gatineau$^{1}$, Armand Jordana$^{1}$, Valentin Guillet$^{2}$, Nicolas Mansard$^{1, 3}$, Olivier Stasse$^{1, 3}$, Philippe Souères$^{1}$
\thanks{$^{\star}$ Equal contribution}
\thanks{$^{1}$ LAAS-CNRS, Université de Toulouse, CNRS, Toulouse, France}
\thanks{$^{2}$ IRT Saint-Exupéry, Toulouse, France}
\thanks{$^{3}$ Artificial and Natural Intelligence Toulouse Institute (ANITI), Toulouse, France}
}

\makeatletter
\providecommand{\bstctlcite}[1]{\@bsphack\@for\@citeb:=#1\do{\edef\@citeb{\expandafter\@firstofone\@citeb}\if@filesw\immediate\write\@auxout{\string\citation{\@citeb}}\fi}\@esphack}
\makeatother

\begin{document}
\bstctlcite{IEEEexample:BSTcontrol}

\makeatletter
\let\@oldmaketitle\@maketitle

\renewcommand{\@maketitle}{\@oldmaketitle
    {\centering
    \setcounter{figure}{0}
    \includegraphics[width=\linewidth]{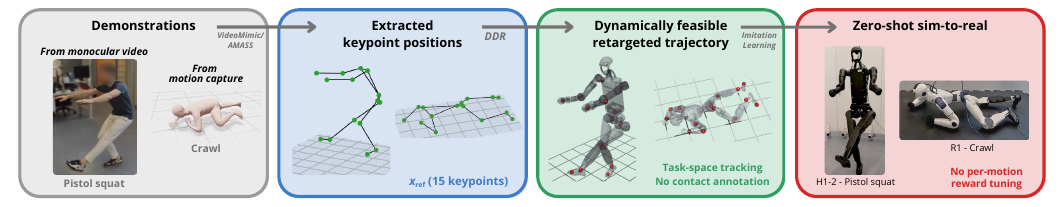}
    \captionof{figure}{The proposed framework uses expert demonstrations from monocular videos, extracts human keypoints using part of the VideoMimic~\cite{videomimic} framework and generates retargeted trajectories using our \textbf{Direct Dynamic Retargeting (DDR)} method. Imitation policies obtained from these trajectories can then be transferred to real humanoid robots.}
    \label{fig:project_overview}
    \par}
    \vspace{-2em}
}
\makeatother

\maketitle
\thispagestyle{empty}
\pagestyle{empty}

\begin{abstract}
Imitation Learning from monocular video demonstrations provides a scalable approach for teaching complex skills to humanoid robots.
However, translating human motion to humanoids requires overcoming significant morphological mismatches.
So far, dynamic retargeting methods obtain physically feasible references by tracking an intermediate geometric retargeting in simulation, usually through its joint angles.
We show that this intermediate step biases the solution through two factors: the space in which the reference is tracked, joint angles rather than task-space keypoints, and the tracked reference itself, the keypoints of the geometric retargeting rather than those of the human demonstration.
We propose Direct Dynamic Retargeting (DDR), which bypasses both: it tracks the human keypoints exclusively in task space with a sampling-based Model Predictive Control solver inside a physics simulator, without contact annotations.
DDR is evaluated in simulation against joint-space dynamic retargeting under controlled conditions (identical solver, budget, and warm start), on three humanoids and on motions from monocular videos and motion capture. The tracking space explains most of the gain in end-effector accuracy and robustness, and tracking the human demonstration directly prevents the transmission of errors from the intermediate geometric reference.
This advantage carries over to Reinforcement Learning imitation policies, which reproduce the human demonstration more faithfully than policies trained on joint-space dynamic references.
These policies are deployed zero-shot on two physical humanoids of very different sizes, H1-2 and R1, including contact-rich motions.
We release the source code and a dataset of dynamically feasible references covering all the flat-ground motions of AMASS on the three humanoids.
Project page: \url{https://gepetto.github.io/robot-motion/ddr/}
\end{abstract}
\section{Introduction} \label{sec:introduction}
Humanoid robots have achieved major breakthroughs in locomotion and loco-manipulation \cite{OmniRetarget}, largely driven by Reinforcement Learning (RL)~\cite{Haarnoja_2019}.
For complex skills, however, reward engineering becomes a bottleneck, and Imitation Learning (IL) techniques such as Behavioral Cloning, DeepMimic~\cite{2018-TOG-deepMimic} and Adversarial Motion Priors \cite{amp, maciej_latent, 2022-TOG-ASE} instead learn from expert demonstrations.
Their performance fundamentally depends on the quality of the references \cite{BeyondMimic, OmniRetarget}, usually obtained with Motion Capture (MoCap) or by a posteriori selection of robot trajectories \cite{Xu_Sheng_Lei_2005, zhang2025hub}.
Monocular videos give access to vast online datasets \cite{videomimic}, but current extraction tools yield noisy and physically inconsistent trajectories.
Moreover, human demonstrations present kinematic (size, limb proportions) and dynamic (mass distribution, inertias, actuation limits) discrepancies with humanoid robots \cite{Bin_Peng_2020}, so they must be retargeted to the robot.

The current prevailing approach, Geometric Retargeting (GR), aims to find the closest robot posture using Inverse Kinematics (IK) \cite{Yang_2023, Bin_Peng_2020, OmniRetarget}, inherently ignoring dynamic constraints.
To address this limitation, Indirect Dynamic Retargeting (IDR) methods track the GR reference within a physics simulator \cite{Dhedin2026-sf, Yoon_2025, pan2025spiderscalablephysicsinformeddexterous}, typically through joint-space targets.
We argue that this intermediate step biases the dynamically feasible solution through two factors: (i) the \emph{tracking space}, since matching the joint angles of the geometric reference anchors the solution to a posture that is not dynamically feasible, and (ii) the \emph{tracked reference}, whose errors become the target of the dynamic stage.

To eliminate these biases, we propose Direct Dynamic Retargeting (DDR), a Model Predictive Control (MPC) framework that computes dynamically feasible trajectories from noisy video data by tracking the human demonstration directly in task space (Fig.~\ref{fig:project_overview}), the geometric solution serving only as a warm start of the solver.
We solve the optimization with the Cross-Entropy Method (CEM) \cite{de2005tutorial} in a physics simulator. Unlike gradient-based solvers \cite{mastalli20crocoddyl}, this sampling approach allows contacts to emerge naturally from task-space tracking without predefined sequences or heuristics.

Under an identical solver, budget and warm start, we show that the tracking space is the dominant factor, increasingly so for dynamic motions, and that the tracked reference matters when the intermediate reference is degraded.
The main contributions are as follows:
\begin{itemize}
    \item \textbf{Direct Dynamic Retargeting}, a CEM-based MPC framework that tracks the human demonstration exclusively in the task space and generates dynamically feasible humanoid trajectories without contact annotations, and applicable to monocular video reconstructions with very few manual choices.
    \item A \textbf{controlled study} on three humanoids (Unitree H1-2, G1 and R1) and two datasets that separates the two factors of the geometric bias. DDR outperforms joint-space dynamic retargeting~\cite{pan2025spiderscalablephysicsinformeddexterous} in end-effector accuracy and robustness, mostly thanks to the tracking space, while tracking the human demonstration prevents the transmission of the errors of the intermediate reference.
    \item Evidence that this advantage \textbf{carries over to closed-loop control}: imitation policies trained on DDR references reproduce the human demonstration more faithfully than those trained on joint-space dynamic references, and are deployed zero-shot on two physical humanoids of very different sizes, H1-2 and R1, including contact-rich motions.
    \item An \textbf{open-source dataset} of dynamically feasible references, released with the source code: all the flat-ground motions of AMASS~\cite{AMASS}, retargeted with DDR on the three humanoids.
\end{itemize}
 \section{Related Work}\label{sec:related_work}
\subsection{Geometric Retargeting}
Transferring human motion to humanoid robots requires overcoming the morphological mismatch between the two embodiments \cite{morph}. 
GR typically relies on Inverse Kinematics to map human keypoints or joint angles to the robot's configuration space \cite{pink, kim2025pyroki}. 
While these methods successfully minimize spatial tracking errors and enforce strict joint limits, they inherently ignore the system's dynamics. 
Consequently, for highly dynamic systems like humanoid robots, purely geometric references frequently result in physically infeasible trajectories that violate balance constraints, torque limits and contact dynamics \cite{opt_moulard}.

\subsection{Indirect Dynamic Retargeting}
To bridge the gap between kinematic similarity and physical validity, IDR approaches introduce a secondary physics-based refinement phase.
Methods in this category follow a two-stage pipeline: an initial GR step generates a kinematic reference, which is then tracked via a joint-space cost within a physics simulator to yield a dynamically feasible trajectory \cite{Dhedin2026-sf, pan2025spiderscalablephysicsinformeddexterous, 2603.09956}.
While ensuring dynamic feasibility, tracking an intermediate geometric reference biases the final motion, both through the joint-space tracking and through the errors of the reference itself.
To the best of our knowledge, no prior work formulates dynamically feasible humanoid retargeting exclusively as a task-space objective, bypassing joint-space tracking entirely and keeping geometric projections solely to warm-start the solver.

\subsection{Imitation Learning from Video Demonstrations}
Imitation learning circumvents reward shaping by reproducing expert data via state tracking \cite{2018-TOG-deepMimic} or adversarial methods \cite{amp, 2022-TOG-ASE, maciej_latent}.
With demonstrations extracted from monocular videos \cite{videomimic}, policies are often trained directly on geometric references, making them highly sensitive to the dynamic infeasibility and kinematic discrepancies of these trajectories \cite{BeyondMimic, OmniRetarget}.
We instead distill dynamically feasible DDR references into closed-loop RL policies for hardware deployment.

\subsection{Motion Datasets for Humanoid Retargeting}
Humanoid datasets derived from human motion databases such as AMASS \cite{AMASS} rely on kinematic mappings \cite{fu2024humanplus}, which yield dynamically infeasible references with sliding contacts and balance issues.
Physics-tracked datasets \cite{wagener2022mocapact} resolve these inconsistencies but are restricted to a single robot morphology.
Our dataset provides DDR references for all the flat-ground motions of AMASS on three humanoids.
 \section{Method} \label{sec:method}
\subsection{Problem Formulation}
\label{sec:problem}
\begin{figure}
    \centering
    \resizebox{\columnwidth}{!}{\begin{tikzpicture}[
    font=\sffamily,
    >=Latex,
title/.style={
        font=\bfseries\Large, 
        text=#1!75!black,
        align=center,
        fill=white,       
        inner sep=6pt     
    },
    space label/.style={
        font=\bfseries\Large, 
        align=left,
        inner sep=7pt
    },
    arrow/.style={
        ->,
        line width=1.2pt,
        draw=darkblue
    },
column/.style={
        rounded corners=5pt,
        draw=#1,
        line width=1.6pt,
        fill=#1!6
    },
    task space/.style={
        rounded corners=4pt,
        draw=taskspaceborder,
        line width=0.8pt,
        fill=taskspace
    },
    joint space/.style={
        rounded corners=4pt,
        draw=jointspaceborder,
        line width=0.8pt,
        dashed,
        fill=jointspace
    },
geometric/.style={
        rounded corners=4pt,
        draw=colGR,
        line width=1pt,
        fill=colGR!30,
        minimum width=4.6cm,  
        minimum height=1.3cm, 
        align=center,
        font=\bfseries\large  
    },
    dynamic/.style={
        rounded corners=4pt,
        draw=colDDR,
        line width=1pt,
        fill=colDDR!30,
        minimum width=4.6cm,
        minimum height=1.3cm,
        align=center,
        font=\bfseries\large  
    },
    dynamic joint/.style={
        rounded corners=4pt,
        draw=colIDR,
        line width=1pt,
        fill=colIDR!30,
        minimum width=4.6cm,
        minimum height=1.3cm,
        align=center,
        font=\bfseries\large  
    },
    output/.style={
        rounded corners=4pt,
        draw=#1,
        line width=1.2pt,
        fill=white,
        minimum width=5.8cm,  
        minimum height=1.4cm,
        align=center,
        font=\large           
    }
]

\node[
    rounded corners=6pt,
    draw=darkblue,
    line width=1.2pt,
    fill=taskspace,
    minimum width=9.5cm,
    minimum height=2.0cm, 
    align=center
] (reference) at (0,8.8) {
    \textbf{\LARGE Human reference}\\[2mm]
    {\huge $x_{\mathrm{ref}}$} \; {\Large (keypoints)} 
};

\node[
    column=colGR,
    minimum width=7.2cm,
    minimum height=7.4cm, 
    anchor=north
] (geo column) at (-8.4,6.0) {};

\node[
    column=colIDR,
    minimum width=7.2cm,
    minimum height=13.8cm, 
    anchor=north
] (ind column) at (0,6.0) {};

\node[
    column=colDDR,
    minimum width=7.2cm,
    minimum height=7.4cm, 
    anchor=north
] (dir column) at (8.4,6.0) {};

\node[
    task space,
    minimum width=6.6cm,
    minimum height=5.6cm,
    anchor=north
] (geo task) at (-8.4,4.8) {}; 

\node[space label, text=taskspaceborder!60!black, anchor=north west] 
    at (geo task.north west) {Task space};

\node[geometric] (geo block) at (-8.4,3.1) 
{
    Geometric\\
    Optimization
};

\node[output=colGR] (geo output) at (-8.4,0.9) 
{
    {\LARGE $x_{\mathrm{GR}}$} \; (keypoints)\\[2.5mm] 
    {\LARGE $Q_{\mathrm{GR}}$} \; (configuration)
};

\draw[arrow] (geo block) -- (geo output);

\node[
    task space,
    minimum width=6.6cm,
    minimum height=5.6cm,
    anchor=north
] (ind task) at (0,4.8) {};

\node[space label, text=taskspaceborder!60!black, anchor=north west] 
    at (ind task.north west) {Task space};

\node[geometric] (ind geo block) at (0,3.1) 
{
    Geometric\\
    Optimization
};

\node[output=colGR] (ind geo output) at (0,0.9) 
{
    {\LARGE $x_{\mathrm{GR}}$} \; (keypoints)\\[2.5mm]
    {\LARGE $Q_{\mathrm{GR}}$} \; (configuration)
};

\draw[arrow] (ind geo block) -- (ind geo output);

\node[
    joint space,
    minimum width=6.6cm,
    minimum height=6.0cm, 
    anchor=north
] (ind joint) at (0,-1.2) {}; 

\node[space label, text=jointspaceborder!80!black, anchor=north west] 
    at (ind joint.north west) {Joint space};

\node[dynamic joint] (ind dynamic) at (0,-2.9) 
{
    Dynamic\\
    Optimization
};

\node[output=colIDR, minimum height=2.4cm] (ind output) at (0,-5.5) 
{
    {\LARGE $x_{\mathrm{IDR}}$} \; (keypoints)\\[2.5mm]
    {\LARGE $U_{\mathrm{IDR}}$} \; (controls)\\[2.5mm]
    {\LARGE $Q_{\mathrm{IDR}}$} \; (configuration)
};

\draw[arrow] (ind dynamic) -- (ind output);

\node[
    task space,
    minimum width=6.6cm,
    minimum height=5.6cm, 
    anchor=north
] (dir task) at (8.4,4.8) {};

\node[space label, text=taskspaceborder!60!black, anchor=north west] 
    at (dir task.north west) {Task space};

\node[dynamic] (dir dynamic) at (8.4,3.1) 
{
    Dynamic\\
    Optimization
};

\node[output=colDDR, minimum height=2.4cm] (dir output) at (8.4,0.7) 
{
    {\LARGE $x_{\mathrm{DDR}}$} \; (keypoints)\\[2.5mm]
    {\LARGE $U_{\mathrm{DDR}}$} \; (controls)\\[2.5mm]
    {\LARGE $Q_{\mathrm{DDR}}$} \; (configuration)
};

\draw[arrow] (dir dynamic) -- (dir output);

\draw[
    darkblue,
    line width=1.2pt
]
(reference.south) -- ++(0,-0.4); 

\draw[
    darkblue,
    line width=1.2pt
]
(-8.4,7.4) -- (8.4,7.4);

\draw[arrow]
    (-8.4,7.4) -- (geo block.north);

\draw[arrow]
    (0,7.4) -- (ind geo block.north);

\draw[arrow]
    (8.4,7.4) -- (dir dynamic.north);

\draw[arrow]
    (ind geo output.south)
    -- (0,-1.2);

\node[title=colGR] (geo title) at (-8.4,6.0)
    {Geometric Retargeting \\ (GR)};

\node[title=colIDR] (ind title) at (0,6.0)
    {Indirect Dynamic \\ Retargeting (IDR)};

\node[title=colDDR] (dir title) at (8.4,6.0)
    {Direct Dynamic \\ Retargeting (DDR)};

\end{tikzpicture}}

    \caption{Retargeting architectures. GR: geometric optimization in task space. IDR: GR followed by a dynamic optimization that tracks it in joint space. DDR: a single dynamic optimization in task space, with respect to the human reference.}
    \label{fig:schematic_GR_IDR_DDR}
\end{figure}

Consider a humanoid robot with a floating base and $n_q$ actuated joints, whose configuration is $q \in SE(3) \times \R^{n_q}$. A trajectory over $T$ timesteps is denoted by $Q = (q_1, \dots, q_T) \in \Q = (SE(3) \times \R^{n_q})^T$, and the corresponding control sequence by $U = (u_0, \dots, u_{T-1}) \in \U = \R^{n_q \times T}$, where $u_t$ are joint position targets tracked by PD controllers.
To characterize physical feasibility, we define a rollout function $S_{s_0}: \U \to \Q$ that integrates the robot dynamics under the controls $U$ from an initial state $s_0$, including the contacts with the environment. In practice, $S_{s_0}$ is a physics simulator. The feasible set contains the trajectories that can be obtained by such a rollout:
\begin{equation}
    \F = \{ Q \in \Q \mid \exists U \in \U,\ S_{s_0}(U) = Q \}.
    \label{eq:feasible_set}
\end{equation}

The motion to reproduce is described in task space by $m$ 3D keypoints. The forward kinematics $fk$ gives the keypoints of a configuration as a matrix $fk(q) \in \R^{m \times 3}$, one row per keypoint, and $FK$ applies it at each timestep, so that a trajectory $Q$ has keypoints $x = FK(Q) = (x_1, \dots, x_T) \in \R^{T \times m \times 3}$. We denote by $\widetilde{\Q} = FK(\Q)$ the kinematically reachable keypoint trajectories and by $\TF = FK(\F)$ the dynamically feasible ones.
The human demonstration provides a keypoint trajectory $x_{\mathrm{ref}} = (x_{\mathrm{ref},1}, \dots, x_{\mathrm{ref},T}) \in \R^{T \times m \times 3}$. Due to human-robot morphological and dynamic differences, $x_{\mathrm{ref}}$ generally lies outside $\widetilde{\Q}$, and by extension, outside $\TF$. Retargeting therefore searches for a robot trajectory whose keypoints are as close as possible to $x_{\mathrm{ref}}$ according to a task-space distance $d$, defined in Sec.~\ref{sec:ddr_cost}.

In this work, $x_{\mathrm{ref}}$ is primarily obtained from monocular videos. The human motion is reconstructed with the SMPL body model~\cite{SMPL:2015} using the VideoMimic pipeline~\cite{videomimic}, and the keypoints are taken at the SMPL joints (Sec.~\ref{sec:exp_setup}). Such reconstructions are noisy and physically inconsistent, and they provide neither contact states nor contact forces. A retargeting method for this setting should therefore not rely on contact annotations, and should limit the propagation of the errors of the reference. The same formulation applies to motion-capture data, which we also use in Sec.~\ref{sec:experiments}.

\subsection{Geometric, Indirect and Direct Retargeting}
\label{sec:approach_overview}
The three families of methods (Fig.~\ref{fig:schematic_GR_IDR_DDR}) differ in the set over which they optimize and in what they track:
\begin{align}
    Q_{\mathrm{GR}} &= \argmin_{Q \in \Q} \, d\big(FK(Q), x_{\mathrm{ref}}\big), \label{eq:gr}\\
    U_{\mathrm{IDR}} &= \argmin_{U \in \U} \, E_q\big(S_{s_0}(U), Q_{\mathrm{GR}}\big), \label{eq:idr}\\
    U_{\mathrm{DDR}} &= \argmin_{U \in \U} \, d\big(FK(S_{s_0}(U)), x_{\mathrm{ref}}\big). \label{eq:ddr}
\end{align}
GR \eqref{eq:gr} optimizes over $\Q$ and ignores the dynamics: its solution $Q_{\mathrm{GR}}$, with keypoints $x_{\mathrm{GR}} = FK(Q_{\mathrm{GR}})$, is kinematically valid but generally not feasible.
IDR \eqref{eq:idr} makes it feasible by tracking $Q_{\mathrm{GR}}$ in the simulator with the joint-space cost $E_q(Q, Q_{\mathrm{GR}}) = \sum_{t=1}^{T} \| q_t \ominus q_{\mathrm{GR},t} \|^2$, where $\ominus$ is the difference in $SE(3) \times \R^{n_q}$. It yields $Q_{\mathrm{IDR}} = S_{s_0}(U_{\mathrm{IDR}}) \in \F$, with keypoints $x_{\mathrm{IDR}} \in \TF$.
DDR \eqref{eq:ddr} removes the intermediate stage and minimizes the task-space distance to the human demonstration over the feasible set, which yields $Q_{\mathrm{DDR}} \in \F$ and $x_{\mathrm{DDR}} \in \TF$.
If \eqref{eq:ddr} were solved to global optimality, then, since $Q_{\mathrm{IDR}} \in \F$,
\begin{equation}
    d(x_{\mathrm{DDR}}, x_{\mathrm{ref}}) = \min_{Q \in \F} d\big(FK(Q), x_{\mathrm{ref}}\big) \le d(x_{\mathrm{IDR}}, x_{\mathrm{ref}}).
\end{equation}
Since the problem is non-convex and the solver sampling-based, this bound is not guaranteed in practice and is evaluated empirically in Sec.~\ref{sec:experiments}.
It nevertheless points to the limitation of two-stage pipelines, which track an intermediate reference $(Q_{\mathrm{GR}}, x_{\mathrm{GR}})$ that generally lies outside $\F$. They can inherit its \emph{geometric bias} through two factors:
(i) the \emph{tracking space}: tracking the joint angles of $Q_{\mathrm{GR}}$ anchors the solution to a posture that is generally not dynamically feasible;
(ii) the \emph{tracked reference}: the kinematic errors of $x_{\mathrm{GR}}$ become the target of the dynamic stage.
DDR avoids both factors: it tracks $x_{\mathrm{ref}}$ in task space, and uses $Q_{\mathrm{GR}}$ only to initialize the solver (Sec.~\ref{sec:ddr_cost}).

\subsection{Task-Space Cost and Solver} \label{sec:ddr_cost}
For the task-space distance, we use the cost of \cite{OmniRetarget}, which combines an absolute position term $E_p$ and a relative shape term $E_l$. For a trajectory $Q$ with keypoints $x = FK(Q)$, we set $d(x, x_{\mathrm{ref}}) = \mathcal{J}(Q, x_{\mathrm{ref}})$, with
\begin{align}
    E_p(Q, x_{\mathrm{ref}}) &= \sum_{t=1}^{T} \sum_{i=1}^{m} \lambda_i \big\| x_{t,i} - x_{\mathrm{ref},t,i} \big\|^2, \label{eq:costs_abs}\\
    E_l(Q, x_{\mathrm{ref}}) &= \sum_{t=1}^{T} \frac{1}{m} \big\| L \big(x_t - x_{\mathrm{ref},t}\big) \big\|^2, \label{eq:costs_lap}\\
    \mathcal{J}(Q, x_{\mathrm{ref}}) &= w_p E_p(Q, x_{\mathrm{ref}}) + w_l E_l(Q, x_{\mathrm{ref}}), \label{eq:cost}
\end{align}
where $x_{t,i}$ and $x_{\mathrm{ref},t,i}$ are the positions of keypoint $i$ at timestep $t$, $\lambda_i$ are per-keypoint weights, and $w_p, w_l$ are scalar weights. The baselines and ablations of Sec.~\ref{sec:experiments} use the same cost with $x_{\mathrm{ref}}$ replaced by the keypoints of a geometric reference, e.g., $\mathcal{J}(Q, x_{\mathrm{GR}})$.
The matrix $L = I - D^{-1}A$ is the random-walk normalized Laplacian of the skeleton graph, with skeletal adjacency matrix $A$ and degree matrix $D$. Applied to the keypoints $x_t$ of a configuration, it gives the position of each keypoint $i$ relative to the centroid of its neighbors $\mathcal{N}(i)$:
\begin{equation}
    (L x_t)_i = x_{t,i} - \frac{1}{D_{ii}} \sum_{j \in \mathcal{N}(i)} x_{t,j}.
\end{equation}
$E_l$ is therefore invariant to a global translation: it constrains the shape of the keypoint configuration but not its position in the world. This position is anchored by $E_p$, which we restrict to the pelvis (Sec.~\ref{sec:exp_setup}), so that the other keypoints are never pulled toward absolute targets that differing limb proportions would make unreachable.

We solve \eqref{eq:ddr} with CEM~\cite{de2005tutorial} in a receding-horizon MPC scheme, using the GPU-accelerated framework of \cite{kurtz2024hydrax}. To reduce the dimension of the search space, the control sequence over the horizon is parameterized by linear splines. At each MPC step, CEM runs $k$ iterations: each iteration samples $N$ control sequences, rolls them out in the simulator, and updates the sampling distribution from the $N_e$ samples of lowest cost, with momentum to prevent a premature collapse of the variance. The sampling distribution is initialized with the geometric solution $Q_{\mathrm{GR}}$ to avoid poor local minima; for DDR, $Q_{\mathrm{GR}}$ is only this warm start and is never tracked.

\subsection{Imitation via Reinforcement Learning} \label{sec:rl_method}
While the retargeted trajectories are dynamically feasible, they are computed offline and rely on open-loop replay. To obtain closed-loop behaviors that recover from initialization errors and disturbances, as required for sim-to-real transfer, we distill each reference into an RL policy, one per motion and robot (Fig.~\ref{fig:project_overview}).
Each imitation task is formulated as a Constrained Markov Decision Process and solved with the Constraints-as-Terminations framework \cite{chane2024cat, roux2025bolt}, built upon the skrl implementation of Proximal Policy Optimization \cite{Schulman2017-mp, serrano2023skrl} in Isaac Lab~\cite{mittal2025isaaclab}.
The observation contains the joint positions and velocities, the previous action, the base linear and angular velocities, the base height, the projected gravity, and a phase variable $\phi_t \in [0,1]$ that parameterizes the reference motion. The action specifies joint position targets, tracked by PD controllers.
The reward follows DeepMimic \cite{2018-TOG-deepMimic} and is identical for all methods: exponential tracking terms on the joint positions and velocities, the root pose and velocity, and the end-effector positions, with penalties on joint accelerations, torques, action rate and foot slip. Its weights are provided with the released code.
Hardware safety constraints limit the joint positions, velocities and torques, and training uses reference state initialization and early termination~\cite{2018-TOG-deepMimic}.
 \section{Evaluation} \label{sec:experiments}
Following Sec.~\ref{sec:approach_overview}, we compare the trajectories of DDR with those of joint-space dynamic retargeting (Sec.~\ref{sec:exp_sota}), ablate the two bias factors (Sec.~\ref{sec:exp_ablation}), and evaluate the imitation policies trained on these references, in simulation and on two physical humanoids (Sec.~\ref{sec:exp_hw}).

\subsection{Experimental Setup} \label{sec:exp_setup}

\paragraph{Robots and keypoints}
The retargeting methods and the policies are evaluated in simulation on three Unitree humanoids of decreasing size: H1-2, G1 and R1. Human motions are represented by the joints of the SMPL model~\cite{SMPL:2015}, from which we use $m=15$ keypoints. Each keypoint is annotated individually in each robot model, at the center of the corresponding joint, except the hands, placed at the end of the arms. The human keypoints are scaled by the ratio of the robot's leg length to the human's leg length~\cite{OmniRetarget}, and no keypoint offset is tuned per robot or per motion.

\paragraph{Datasets}
Two datasets with complementary difficulties are used. The Monocular-Video dataset contains six motions (squat, kung-fu, one-foot balance, pistol squat, balancing stick and walk), each performed by three subjects and reconstructed from monocular videos (Sec.~\ref{sec:problem}). They contain visual noise and physical inconsistencies, and their foot contacts are manually annotated for evaluation only: no method uses them. The ACCAD subset of the AMASS motion-capture dataset~\cite{AMASS} is used with 236 clips, without reconstruction noise and including fast and highly dynamic motions.

\paragraph{Baselines}
GR solves \eqref{eq:gr} with the cost $\mathcal{J}$ of \eqref{eq:cost}. Being dynamically infeasible, it serves as an error lower bound and the warm start for dynamic methods, rather than a direct competitor. 
IDR-J is our implementation of SPIDER~\cite{pan2025spiderscalablephysicsinformeddexterous}: it solves \eqref{eq:idr}, i.e., it tracks the joint angles and the pelvis pose of GR.
IDR-JT minimizes $w_q E_q(S_{s_0}(U), Q_{\mathrm{GR}}) + \mathcal{J}(S_{s_0}(U), x_{\mathrm{GR}})$: it adds our task-space cost, computed on the keypoints of GR, to IDR-J. This hybrid is close in spirit to DynaRetarget~\cite{Dhedin2026-sf}, which combines joint-space and task-space terms, but is not a reimplementation of it.
Both IDR baselines therefore track GR in joint space, whereas DDR minimizes $\mathcal{J}(S_{s_0}(U), x_{\mathrm{ref}})$.

\paragraph{Solver settings}
All dynamic methods run CEM-based MPC at $50$\,Hz in MuJoCo~\cite{todorov2012mujoco} with identical budgets, 4-knot linear splines, and geometric warm starts, differing only in their tracked costs. The horizon is hand-tuned for each Monocular-Video motion, between $0.6$ and $1.4$\,s, and is $0.8$\,s for ACCAD. All other hyperparameters are shared: $N=1024$ samples, $k=5$ iterations per MPC step, $N_e=10$ elites, initial and minimum noise $\sigma_{\text{start}}=0.25$ and $\sigma_{\text{min}}=0.01$, and momentums $\alpha_{\text{mean}}=0.95$ and $\alpha_{\text{cov}}=0.2$. The optimization ends when the horizon reaches the end of the clip. The cost weights are also identical for all motions, datasets and robots: $w_p=1$ with $E_p$ restricted to the pelvis ($\lambda_{\text{pelvis}}=1$, $\lambda_i=0$ otherwise), $w_l=2$ and $w_q=1$. Each method is run with 5 seeds on Monocular-Video (90 runs per robot) and one seed on ACCAD (708 runs over the three robots). On one H100 GPU, a retargeting takes about $3$\,min per second of motion on H1-2 and $45$\,s on G1 and R1.

\paragraph{Metrics}
Spatial fidelity is measured with respect to the human reference $x_{\mathrm{ref}}$, in mm. The Laplacian error, computed with the operator of \eqref{eq:costs_lap}, measures the local shape of the keypoint configuration, and the end-effector error is the position error of the hands and feet. A run is successful if the maximum deviation of its pelvis from the reference stays below a threshold. The contact mismatch is the percentage of frames in which the simulated foot contact differs from the manual annotation. A simulated foot is considered in contact if it is below $5$\,cm with a horizontal speed under $0.5$\,m/s. Results are reported as mean $\pm$ standard deviation over clips, or over seeds for the policies.

\subsection{Comparison with Joint-Space Dynamic Retargeting}
\label{sec:exp_sota}

\paragraph{Retargeting accuracy}
Table~\ref{tab:accuracy} reports the errors of the retargeted trajectories with respect to the human reference. GR has the lowest errors, since it ignores the dynamics. Among the dynamically feasible methods, DDR is the most accurate in 11 of the 12 robot--dataset--metric combinations. On average over the robots, it reduces the end-effector error of IDR-J by 22\% on Monocular-Video and by 43\% on ACCAD. IDR-JT narrows the gap in Laplacian error, which stays within 5\% of that of DDR, but its end-effector error remains higher, especially on ACCAD. Adding a task-space term to a joint-space cost is thus not enough to match a task-space cost alone.

\begin{table*}
\centering
\caption{Laplacian and end-effector errors (mm, $\downarrow$) of the retargeted trajectories with respect to the human reference, mean $\pm$ standard deviation over clips, on the Monocular-Video and ACCAD datasets.
The best results among dynamically feasible methods are in \textbf{bold}. The geometric baseline is \underline{underlined}.}
\label{tab:accuracy}
\resizebox{\linewidth}{!}{\begin{tabular}{l|cccc|cccc}
\toprule
 & \multicolumn{4}{c}{Monocular-Video} & \multicolumn{4}{c}{ACCAD} \\
Robot & GR & IDR-J & IDR-JT & DDR (ours) & GR & IDR-J & IDR-JT & DDR (ours) \\
\midrule
\multicolumn{9}{c}{\textit{Laplacian error [mm]} ($\downarrow$)} \\
H1-2 & \underline{$37.1{\pm}1.5$} & $66.4{\pm}17.3$ & $52.9{\pm}15.9$ & $\mathbf{52.7{\pm}11.8}$ & \underline{$41.5{\pm}8.7$} & $90.9{\pm}36.9$ & $66.7{\pm}24.3$ & $\mathbf{65.6{\pm}21.1}$ \\
G1 & \underline{$29.1{\pm}1.8$} & $43.2{\pm}11.0$ & $37.4{\pm}7.6$ & $\mathbf{37.1{\pm}5.8}$ & \underline{$31.4{\pm}3.7$} & $64.2{\pm}23.7$ & $50.4{\pm}16.7$ & $\mathbf{48.2{\pm}13.6}$ \\
R1 & \underline{$21.2{\pm}1.3$} & $41.8{\pm}8.0$ & $\mathbf{36.3{\pm}6.8}$ & $37.6{\pm}5.4$ & \underline{$24.3{\pm}4.3$} & $60.0{\pm}28.4$ & $49.9{\pm}23.9$ & $\mathbf{47.6{\pm}21.8}$ \\
\midrule
\multicolumn{9}{c}{\textit{End-effector error [mm]} ($\downarrow$)} \\
H1-2 & \underline{$52.0{\pm}5.2$} & $101.9{\pm}47.8$ & $79.9{\pm}41.7$ & $\mathbf{76.9{\pm}42.0}$ & \underline{$63.0{\pm}15.2$} & $139.9{\pm}75.0$ & $99.2{\pm}56.3$ & $\mathbf{79.9{\pm}42.1}$ \\
G1 & \underline{$29.7{\pm}4.8$} & $65.9{\pm}40.4$ & $53.9{\pm}37.9$ & $\mathbf{49.1{\pm}23.5}$ & \underline{$38.3{\pm}11.5$} & $111.2{\pm}74.5$ & $79.3{\pm}54.9$ & $\mathbf{58.4{\pm}34.9}$ \\
R1 & \underline{$37.4{\pm}3.4$} & $72.0{\pm}35.7$ & $63.4{\pm}38.5$ & $\mathbf{60.7{\pm}24.5}$ & \underline{$35.2{\pm}7.8$} & $108.2{\pm}73.2$ & $88.8{\pm}60.0$ & $\mathbf{67.7{\pm}40.6}$ \\
\bottomrule
\end{tabular}}
\end{table*}

\begin{figure}
    \centering
    \includegraphics[width=\linewidth, trim=8 7 0 22, clip]{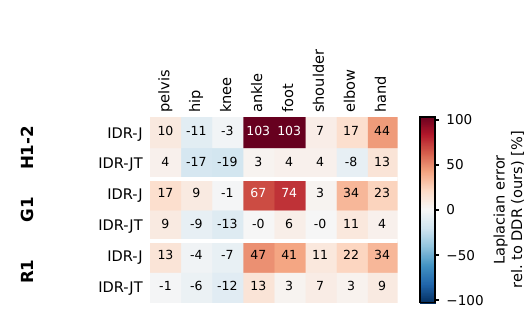}
    \caption{Per-keypoint Laplacian error of IDR-J and IDR-JT relative to DDR (\%), averaged over clips. Red: higher error than DDR; blue: lower error than DDR.}
    \label{fig:keypoints}
\end{figure}

\paragraph{Per-keypoint accuracy}
Fig.~\ref{fig:keypoints} locates this difference. Joint-space tracking mainly degrades the extremities. On all three robots, the error of IDR-J at the ankles and feet is much higher than that of DDR (by about 100\% on H1-2, 70\% on G1 and 45\% on R1), while it stays close to DDR at the hips and knees. IDR-JT shows a trade-off: it is more accurate than DDR at the hips and knees, but not at the ankles, feet and hands. This is why it matches DDR in Laplacian error but not in end-effector error.
Two effects explain this pattern. First, matching joint angles is not equivalent to matching keypoints: a small angle error at the hip or the knee is amplified along the kinematic chain into a larger displacement of the foot. Second, the end-effectors carry contact and balance: they are where the dynamics most require departing from the geometric posture, and where a term that pulls toward this posture costs most. The hips and knees, in contrast, are well constrained by the geometric reference, and anchoring them to it helps.

\paragraph{Robustness}
DDR is also more robust (Table~\ref{tab:success_compact}). On ACCAD, the share of runs whose pelvis stays within 10\,cm of the reference rises from 31\% for IDR-J to 56\% for DDR. The gap is smaller on the slower Monocular-Video motions, but persists. Most failures are deviations rather than falls: on clips too fast for the robot, the solver can no longer follow the keypoints and the pelvis drifts while the robot keeps its balance. A joint-space term makes such drifts more frequent, since the solver trades it against balance when the geometric posture cannot be reached at the speed of the motion.

\paragraph{Contacts}
Contacts, in contrast, do not discriminate between the dynamic methods (Table~\ref{tab:contact_universal}). All of them reduce the contact mismatch of GR, from 18\% to less than 10\% of the frames on average over the robots, mostly on R1. The simulator only admits physical contacts, whatever the tracking space, which affects where the feet are placed but not whether their contacts are physical. Monocular reconstructions provide no contact information, and DDR reaches the contact quality of the baselines without any contact annotation.

\begin{table}[]
\centering
\footnotesize
\setlength{\tabcolsep}{3pt}
\caption{Success rate [\%] ($\uparrow$): runs whose maximum pelvis deviation stays below the threshold, averaged over the three robots. Bold: best dynamic method.}
\label{tab:success_compact}
\begin{tabular}{llcccc}
\toprule
 & & \multicolumn{4}{c}{Pelvis deviation threshold} \\
\cmidrule(lr){3-6}
Dataset & Method & $<$10\,cm & $<$20\,cm & $<$30\,cm & $<$50\,cm \\
\midrule
Monocular-Video & IDR-J & 63.0 & 91.5 & 98.5 & 99.6 \\
 & IDR-JT & 64.4 & 91.5 & 98.1 & \textbf{100.0} \\
 & DDR (ours) & \textbf{77.7} & \textbf{97.3} & \textbf{99.6} & \textbf{100.0} \\
\midrule
ACCAD & IDR-J & 31.1 & 57.3 & 74.0 & 87.9 \\
 & IDR-JT & 34.5 & 65.0 & 78.7 & 89.8 \\
 & DDR (ours) & \textbf{55.8} & \textbf{79.1} & \textbf{88.7} & \textbf{95.5} \\
\bottomrule
\end{tabular}
\end{table}

\begin{table}[]
\centering
\footnotesize
\setlength{\tabcolsep}{3pt}
\caption{Contact mismatch with manual annotations on Monocular-Video (\% of frames, mean $\pm$ standard deviation over clips; $\downarrow$).}
\label{tab:contact_universal}
\begin{tabular}{lcccc}
\toprule
Method & H1-2 & G1 & R1 & All robots \\
\midrule
GR & $11.2{\pm}15.8$ & $9.3{\pm}13.7$ & $33.2{\pm}20.6$ & $17.9{\pm}19.9$ \\
IDR-J & $9.6{\pm}10.2$ & $8.8{\pm}9.3$ & $9.9{\pm}9.1$ & $9.4{\pm}9.4$ \\
IDR-JT & $7.3{\pm}8.9$ & $7.9{\pm}9.0$ & $9.0{\pm}9.2$ & $8.1{\pm}8.9$ \\
DDR (ours) & $7.5{\pm}9.2$ & $7.7{\pm}8.8$ & $9.3{\pm}8.8$ & $8.1{\pm}8.8$ \\
\bottomrule
\end{tabular}\end{table}

\subsection{Ablation of the Tracking Space and the Tracked Reference}
\label{sec:exp_ablation}
DDR differs from the IDR baselines in both factors of Sec.~\ref{sec:approach_overview}. To separate them, we introduce two ablations, with the same solver, budget and warm start:
\begin{itemize}
    \item DDR-G minimizes $\mathcal{J}(S_{s_0}(U), x_{\mathrm{GR}})$. It tracks the same reference as IDR-J and IDR-JT, but in task space only. From IDR-J to IDR-JT and DDR-G, only the tracking space (i) changes. From DDR-G to DDR, only the tracked reference (ii) changes.
    \item DDR-GS minimizes $\mathcal{J}(S_{s_0}(U), x_{\mathrm{GR\text{-}S}})$, where $x_{\mathrm{GR\text{-}S}}$ are the keypoints of a degraded geometric reference GR-S. GR-S adds a smoothness term on the joint trajectory to \eqref{eq:gr} and removes the joint limits. DDR-GS measures how the errors of the tracked reference propagate (ii).
\end{itemize}

\paragraph{Isolating the tracking space}
IDR-J, IDR-JT and DDR-G track the same reference and differ only in the tracking space. DDR-G is already almost as accurate as DDR and clearly more accurate than IDR-J: its end-effector error is up to 49\% lower than that of IDR-J, against 48\% for DDR. Removing the joint-space term from IDR-JT, which yields DDR-G, also reduces the end-effector error, by up to 29\%. DDR-G and DDR are moreover within 2\,mm of each other on every keypoint, so the per-keypoint pattern of Fig.~\ref{fig:keypoints} also comes from the tracking space.

\begin{figure}
    \centering
    \includegraphics[width=\linewidth]{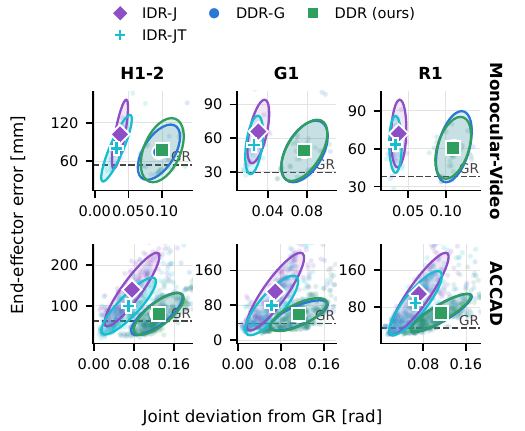}
    \caption{Tracking space (i). End-effector error versus joint deviation from the GR reference, per clip. Dots: clips; markers: means; ellipses: 1.5 std; dashed: median GR error.}
    \label{fig:cross}
\end{figure}

\begin{figure}
    \centering
    \includegraphics[width=\linewidth]{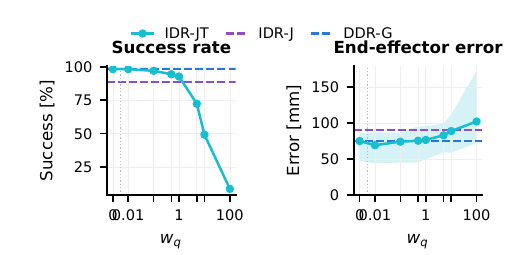}
    \caption{Tracking space (i). Success rate (left) and end-effector error (right) versus the joint-tracking weight $w_q$ for H1-2 on ACCAD. Lines: median over clips; bands: interquartile range; horizontal lines: DDR-G, which corresponds to $w_q=0$.}
    \label{fig:sweep}
\end{figure}

\paragraph{Weight of the joint-space term}
Fig.~\ref{fig:cross} shows that this relation holds clip by clip: the closer a method stays to the joint angles of GR, the larger its end-effector error. To verify that the joint-space term causes this degradation, Fig.~\ref{fig:sweep} varies its weight $w_q$ in IDR-JT on H1-2 with ACCAD, from $w_q=0$, which corresponds to DDR-G, to $w_q=100$. All points of this sweep track the same reference. As $w_q$ grows, the share of runs whose pelvis deviates by more than 50\,cm rises from 1.7\% to 91.5\%, and the median end-effector error grows from 75 to 102\,mm. The more weight is given to the joint angles of GR, the less accurate and robust the retargeting.

\begin{figure}
    \centering
    \includegraphics[width=\linewidth]{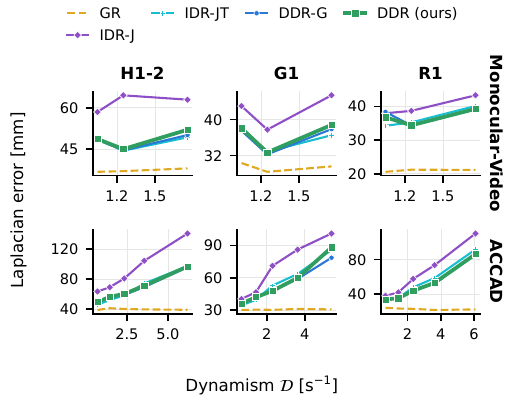}
    \caption{Tracking space (i). Laplacian error versus motion dynamism $\mathcal{D}$ (median per quantile bin: 3 on Monocular-Video, 5 on ACCAD).}
    \label{fig:dynamism}
\end{figure}

\paragraph{Effect of motion dynamism}
We define the dynamism $\mathcal{D}$ [s$^{-1}$] of a clip as the 95th percentile of the joint speeds of the human reference, normalized by the leg length. On ACCAD, the gap between joint-space and task-space tracking widens with $\mathcal{D}$ (Fig.~\ref{fig:dynamism}). In the most dynamic class, the median error of IDR-J exceeds that of DDR by up to 46\% in Laplacian error and 98\% in end-effector error. IDR-JT stays close to DDR in Laplacian error (within 11\% at every level), but its end-effector error is more than 30\% higher in this class, and competitive with DDR on the slowest motions only. Fast motions require larger departures from the geometric posture to preserve balance and contact. The cost of the joint-space term therefore grows with the dynamism of the motion.

\begin{figure}
    \centering
    \includegraphics[width=\linewidth]{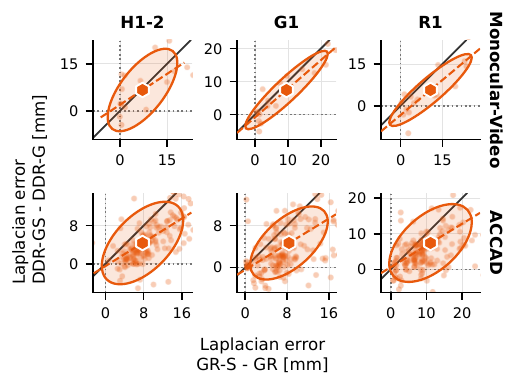}
    \caption{Tracked reference (ii). Propagation of the reference error: Laplacian error of DDR-GS over DDR-G versus error of GR-S over GR, per clip. Solid: $y=x$; dashed: linear fit.}
    \label{fig:delta}
\end{figure}

\paragraph{Tracked reference}
DDR-G and DDR differ only in the tracked reference (ii). With the geometric references of Table~\ref{tab:accuracy}, their errors differ by at most 3.2\,mm, far below the dispersion over clips, and their success rates are the same. The tracked reference thus has little effect when the geometric reference is accurate. It matters when this reference is degraded. A method that tracks it cannot correct its errors, since it has no access to the human motion. Fig.~\ref{fig:delta} shows that DDR-GS inherits a large share of the degradation of GR-S in Laplacian error: the slope of the linear fit is at least 0.6 in every panel, and up to 0.87. DDR is not exposed to this propagation: its cost depends only on $x_{\mathrm{ref}}$, and $Q_{\mathrm{GR}}$ only serves as the warm start.

\subsection{Closed-Loop Imitation}
\label{sec:exp_hw}
The retargeted trajectories are open-loop references. We finally verify that their accuracy is preserved by the imitation policies trained on them, in simulation and then zero-shot on two physical humanoids.

\paragraph{Policies in simulation}
We train policies (Sec.~\ref{sec:rl_method}) on the GR, IDR-J and DDR references of the squat, kung-fu and walk of Monocular-Video, on the three robots (Table~\ref{tab:rollout_human}). GR is included because policy training may compensate for part of its dynamic infeasibility. The final training return is computed with respect to the reference each policy imitates, so it is not comparable across methods, and the rollouts are compared with the human reference instead. DDR policies are the most accurate in 14 of the 18 robot--motion--metric combinations, including where GR reaches a higher return, and more accurate than IDR-J policies in 16 (within one standard deviation on the two exceptions, both on the G1 squat). On H1-2, the end-effector error of the DDR policies is about half that of the IDR-J policies and a quarter lower than that of the GR policies. Although the IDR-J references are dynamically feasible, their policies thus inherit their bias: a policy reproduces the reference it imitates, so the accuracy gained at the retargeting stage carries over to closed loop.

\begin{table*}
\centering
\caption{Imitation policies: final training return, computed with respect to the reference each policy imitates and therefore not comparable across methods, and rollout errors with respect to the human reference (mm, $\downarrow$). Mean $\pm$ standard deviation over seeds (for the errors, over the seeds whose policy completes the first cycle). Best error in \textbf{bold}.}
\label{tab:rollout_human}
\resizebox{\linewidth}{!}{\begin{tabular}{l|ccc|ccc|ccc}
\toprule
 & \multicolumn{3}{c}{Squat} & \multicolumn{3}{c}{Kung-fu} & \multicolumn{3}{c}{Walk} \\
Method & H1-2 & G1 & R1 & H1-2 & G1 & R1 & H1-2 & G1 & R1 \\
\midrule
\multicolumn{10}{c}{\textit{Final training return}} \\
GR & $357.9{\pm}5.5$ & $414.8{\pm}3.9$ & $318.7{\pm}3.8$ & $349.1{\pm}15.0$ & $401.9{\pm}4.1$ & $345.2{\pm}13.5$ & $79.8{\pm}0.2$ & $87.8{\pm}0.3$ & $78.8{\pm}0.4$ \\
IDR-J & $328.4{\pm}16.2$ & $433.7{\pm}4.3$ & $321.5{\pm}2.6$ & $387.9{\pm}26.2$ & $341.3{\pm}172.4$ & $338.8{\pm}15.1$ & $70.4{\pm}2.7$ & $87.2{\pm}2.0$ & $75.3{\pm}1.3$ \\
DDR (ours) & $377.6{\pm}10.4$ & $442.1{\pm}4.0$ & $326.2{\pm}4.0$ & $395.2{\pm}15.3$ & $418.0{\pm}12.7$ & $338.7{\pm}8.4$ & $85.5{\pm}0.2$ & $86.2{\pm}1.0$ & $77.7{\pm}1.0$ \\
\midrule
\multicolumn{10}{c}{\textit{Laplacian error [mm]} ($\downarrow$)} \\
GR & $69.8{\pm}9.8$ & $55.5{\pm}7.0$ & $62.3{\pm}5.7$ & $82.2{\pm}12.9$ & $\mathbf{53.0{\pm}3.7}$ & $55.4{\pm}7.4$ & $53.7{\pm}3.4$ & $36.7{\pm}1.7$ & $42.8{\pm}2.5$ \\
IDR-J & $71.5{\pm}7.0$ & $\mathbf{49.4{\pm}7.7}$ & $58.9{\pm}3.6$ & $82.2{\pm}6.3$ & $59.7{\pm}8.7$ & $55.3{\pm}5.8$ & $68.7{\pm}3.2$ & $43.1{\pm}5.6$ & $44.3{\pm}3.8$ \\
DDR (ours) & $\mathbf{61.3{\pm}2.8}$ & $52.4{\pm}8.5$ & $\mathbf{58.1{\pm}5.8}$ & $\mathbf{71.6{\pm}4.9}$ & $58.3{\pm}6.1$ & $\mathbf{51.5{\pm}1.2}$ & $\mathbf{50.3{\pm}2.3}$ & $\mathbf{36.4{\pm}1.3}$ & $\mathbf{39.2{\pm}2.0}$ \\
\midrule
\multicolumn{10}{c}{\textit{End-effector error [mm]} ($\downarrow$)} \\
GR & $77.9{\pm}14.8$ & $55.1{\pm}5.6$ & $79.9{\pm}6.1$ & $99.0{\pm}20.4$ & $\mathbf{63.7{\pm}5.2}$ & $73.6{\pm}9.7$ & $98.2{\pm}56.6$ & $68.5{\pm}9.6$ & $123.5{\pm}7.6$ \\
IDR-J & $115.1{\pm}58.2$ & $\mathbf{52.0{\pm}5.2}$ & $79.4{\pm}5.0$ & $137.3{\pm}22.9$ & $78.2{\pm}10.9$ & $74.6{\pm}7.6$ & $137.8{\pm}21.8$ & $77.3{\pm}20.3$ & $150.7{\pm}38.1$ \\
DDR (ours) & $\mathbf{49.7{\pm}3.5}$ & $53.5{\pm}10.2$ & $\mathbf{76.3{\pm}5.6}$ & $\mathbf{80.5{\pm}5.7}$ & $75.4{\pm}7.5$ & $\mathbf{71.6{\pm}3.8}$ & $\mathbf{75.9{\pm}23.2}$ & $\mathbf{45.2{\pm}7.8}$ & $\mathbf{103.5{\pm}14.9}$ \\
\bottomrule
\end{tabular}}
\end{table*}

\begin{figure}
    \centering
    \includegraphics[width=0.9\linewidth, trim=6 7 6 6, clip]{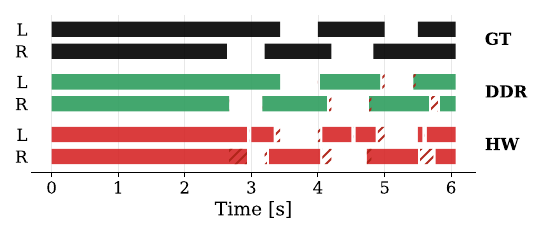}
    \caption{Sim-to-real gap on the H1-2 walk: left/right foot contacts of the annotated ground truth (GT), the DDR simulation and the hardware (HW). Hatched: mismatches with GT.}
    \label{fig:contacts_hw}
\end{figure}

\begin{figure}[]
    \centering
    \begin{subfigure}[t]{0.24\linewidth}
        \centering
        \includegraphics[width=\linewidth]{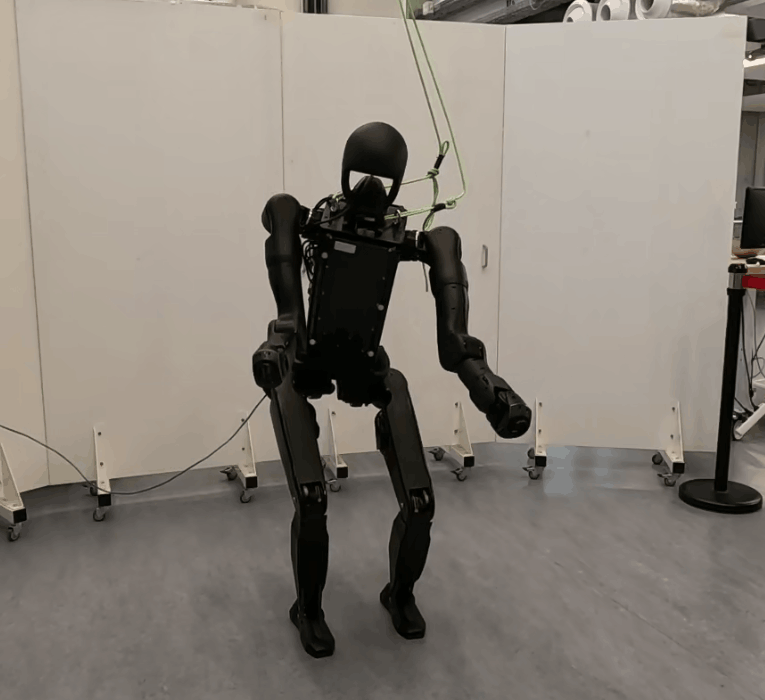}
        \caption{Squat}
    \end{subfigure}\hfill
    \begin{subfigure}[t]{0.24\linewidth}
        \centering
        \includegraphics[width=\linewidth]{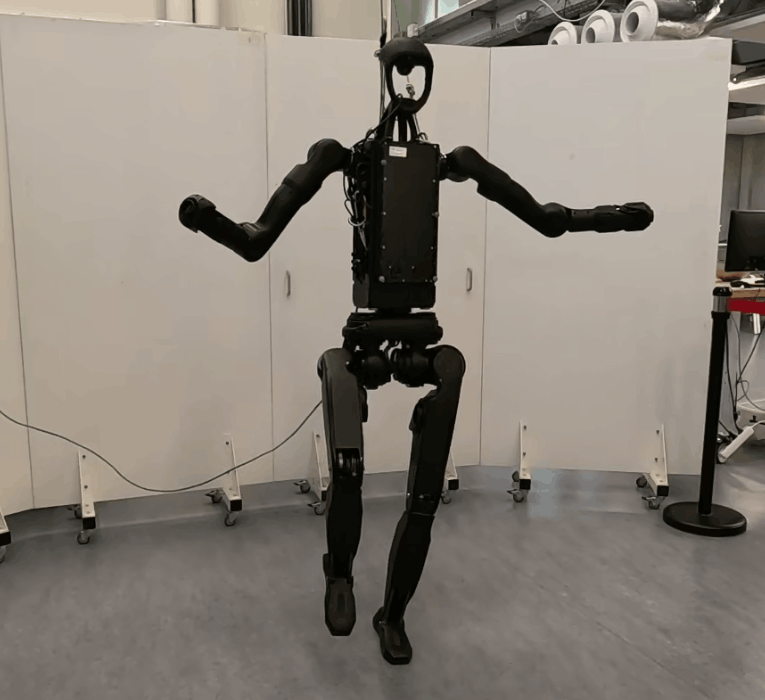}
        \caption{Kung-fu}
    \end{subfigure}\hfill
    \begin{subfigure}[t]{0.24\linewidth}
        \centering
        \includegraphics[width=\linewidth]{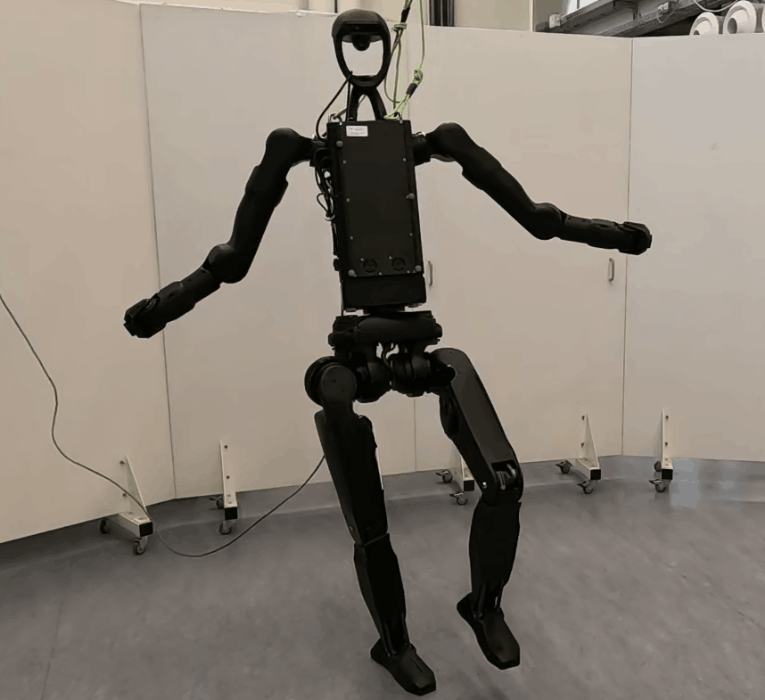}
        \caption{\centering One-foot balance}
    \end{subfigure}\hfill
    \begin{subfigure}[t]{0.24\linewidth}
        \centering
        \includegraphics[width=\linewidth]{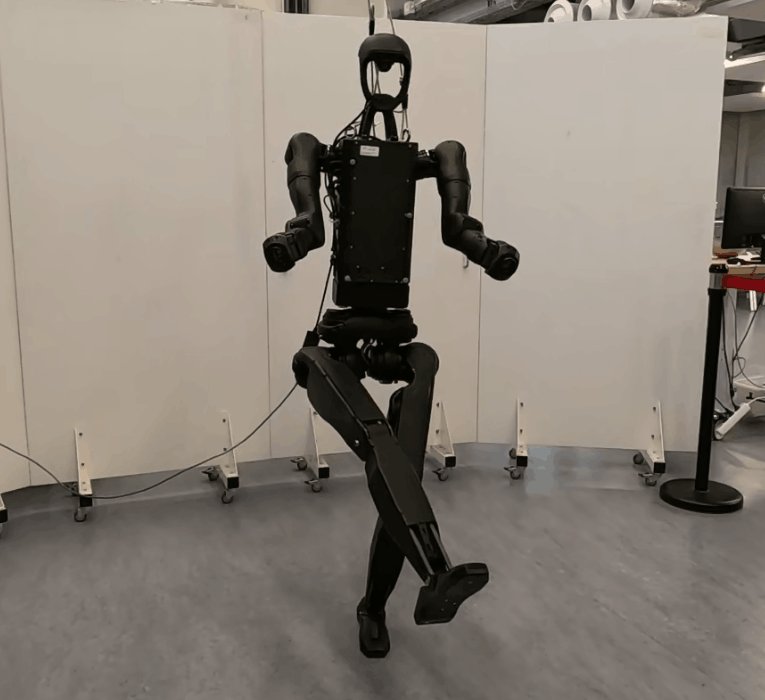}
        \caption{Pistol squat}
    \end{subfigure}
    \\
    \begin{subfigure}[t]{0.24\linewidth}
        \centering
        \includegraphics[width=\linewidth]{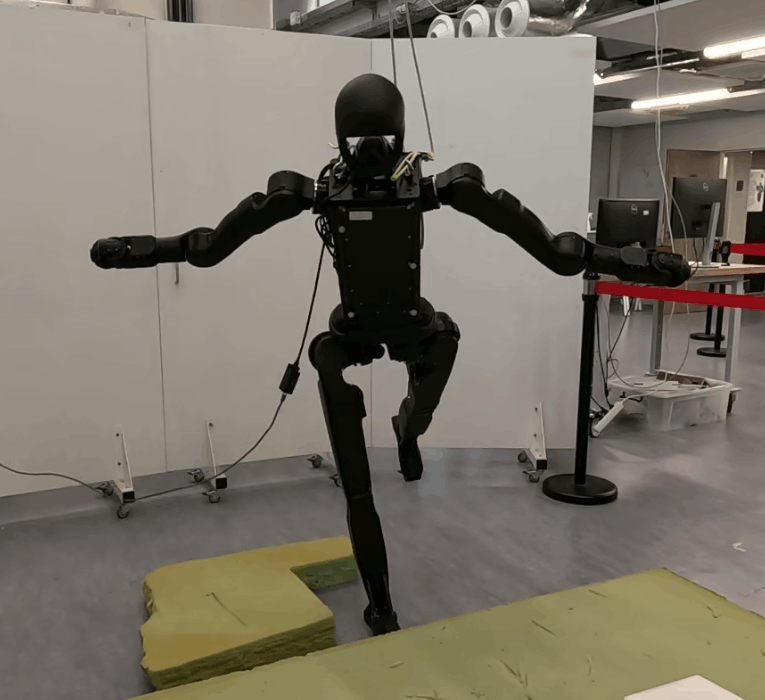}
        \caption{Balancing stick}
    \end{subfigure}\hfill
    \begin{subfigure}[t]{0.24\linewidth}
        \centering
        \includegraphics[width=\linewidth]{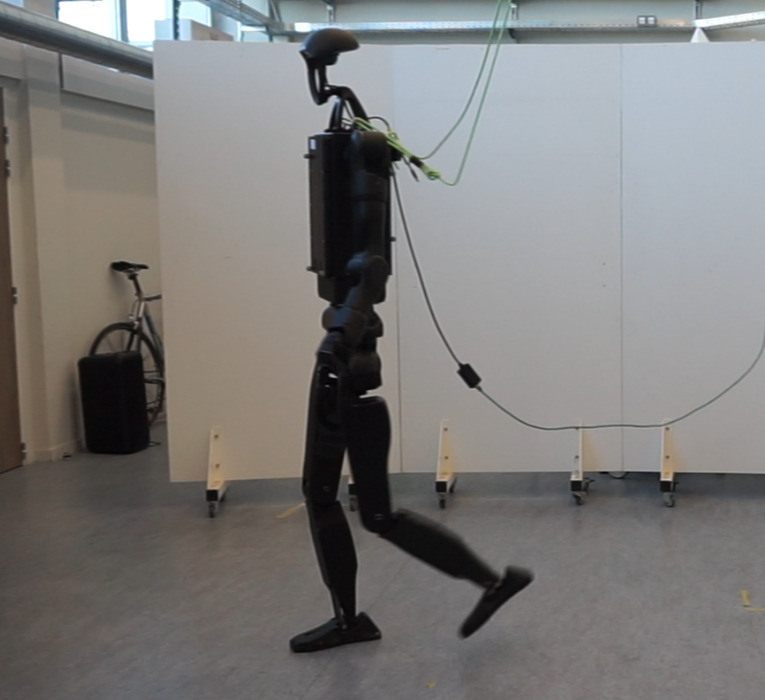}
        \caption{Walk}
    \end{subfigure}\hfill
    \begin{subfigure}[t]{0.24\linewidth}
        \centering
        \includegraphics[width=\linewidth]{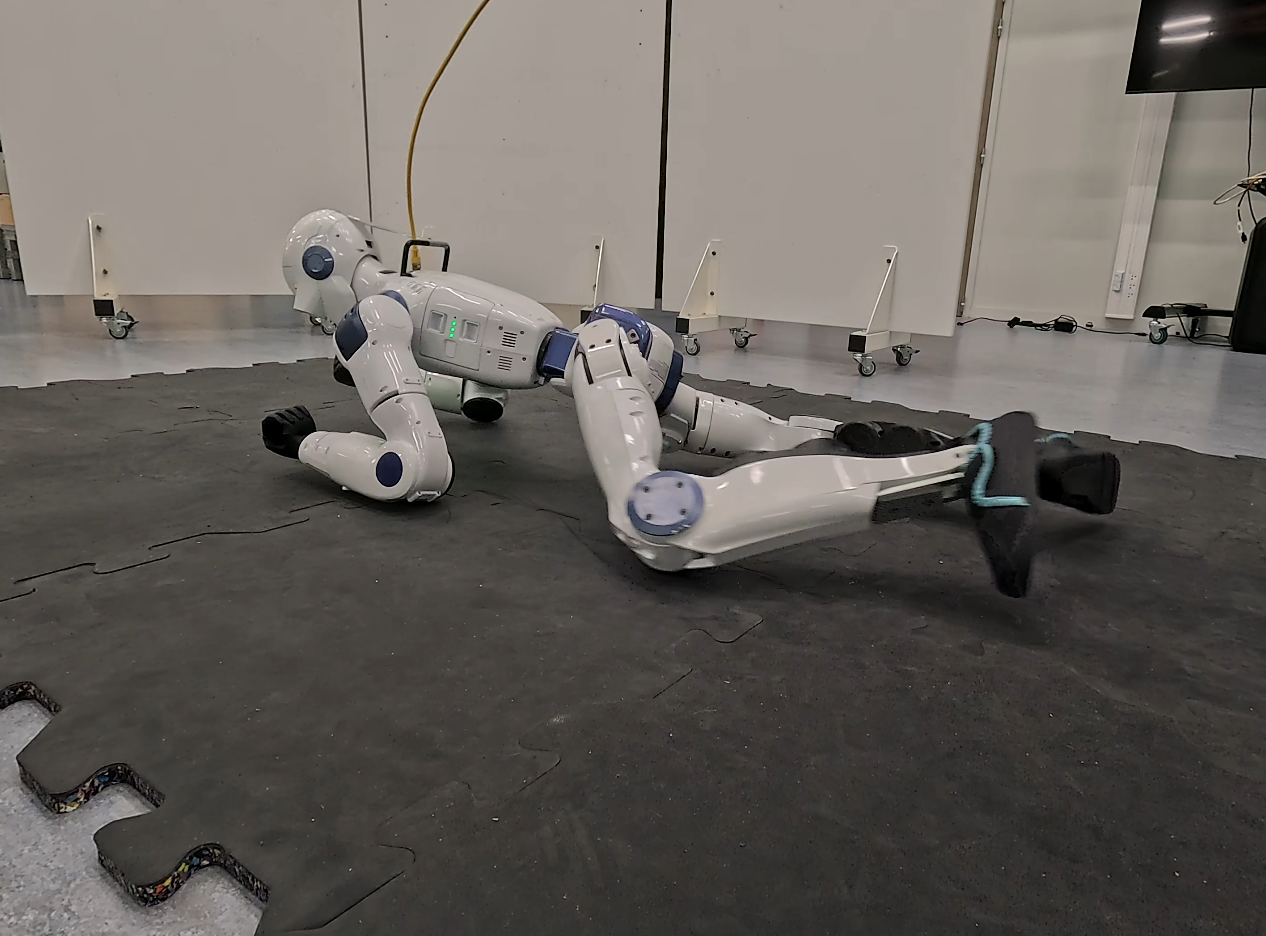}
        \caption{Crawl}
    \end{subfigure}\hfill
    \begin{subfigure}[t]{0.24\linewidth}
        \centering
        \includegraphics[width=\linewidth]{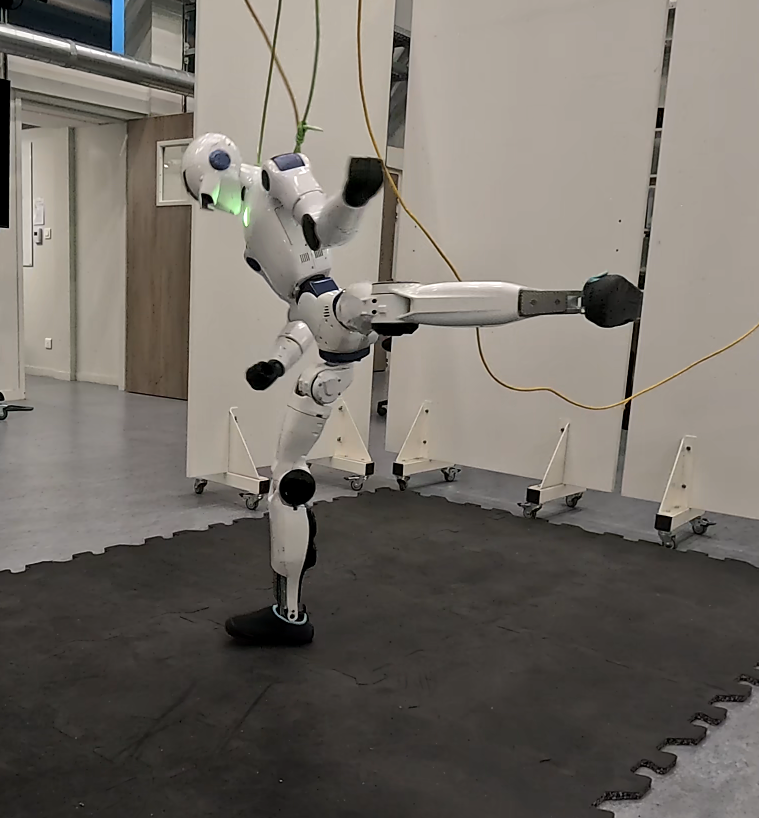}
        \caption{Side kick}
    \end{subfigure}
    \caption{Zero-shot hardware deployment of DDR policies: (a)--(f) H1-2 on the Monocular-Video motions, (g)--(h) R1 on the ACCAD crawl and side kick.}
    \label{fig:deployed_motions}
\end{figure}

\paragraph{Deployment on H1-2}
DDR policies are deployed zero-shot, without per-motion tuning (Fig.~\ref{fig:deployed_motions} and supplementary video). The human-sized H1-2 uses proprioception and motion capture for state estimation. It performs the six motions reconstructed from monocular videos: squat, kung-fu, one-foot balance, pistol squat, balancing stick and walk. For these motions, the whole pipeline, from a monocular video to a policy on the real robot, runs without contact annotation and with the same cost weights. We quantify the sim-to-real gap on the walk of Table~\ref{tab:rollout_human}, with the contact mismatch of Table~\ref{tab:contact_universal}, a metric that depends on the foot motion and thus reflects the quality of the gait (Fig.~\ref{fig:contacts_hw}). Over one hardware run, the contact mismatch with the annotated ground truth is 3.7\% in simulation and 9.4\% on the real robot, of the same order as that of the retargeted trajectories themselves (Table~\ref{tab:contact_universal}). The contact pattern of the human gait is thus largely preserved on the real robot.

\paragraph{Deployment on R1}
The smaller R1 relies on proprioception only. It performs the Monocular-Video squat (supplementary video), and two ACCAD motions that involve multiple contacts or a single-leg support and would be difficult to specify with hand-designed rewards: a crawl supported by the knees and forearms, and a side kick balanced on one leg (Fig.~\ref{fig:deployed_motions}). Their references are retargeted with the same cost and weights as all other motions, and the policies use the same reward (Sec.~\ref{sec:rl_method}). The contact sequences of the two ACCAD motions are specified nowhere in the pipeline: they result from tracking the human keypoints in simulation, and the policies reproduce them on the real robot. Together with the H1-2, this shows that the pipeline transfers zero-shot, with the same cost weights, to robots of very different sizes. The motions of the full-size H1-2 are less clean than those of the R1, in line with the common observation that larger robots are harder to control.

\newcommand{\imgplaceholder}[1]{\fcolorbox{red}{white}{\parbox[c][0.9\linewidth][c]{0.85\linewidth}{\centering\scriptsize\color{red}#1}}}
 \section{Conclusion} \label{sec:conclusion}
We introduced Direct Dynamic Retargeting (DDR), which tracks the human demonstration exclusively in task space with a physics-in-the-loop solver.
Since it requires neither contact annotations nor a per-robot tuning of its cost, DDR applies, with very few manual choices, to noisy human motions reconstructed from monocular videos, as well as to motion-capture data, with the same weights for all robots and datasets.
The tracking space is the dominant of the two factors through which two-stage pipelines bias dynamic retargeting. Joint-space tracking constrains the end-effectors, which carry contact and balance, precisely where the dynamics require departing from the geometric posture. This degrades accuracy and robustness, all the more as the motion is dynamic. The tracked reference matters when the geometric reference is degraded: more than half of its error is then transmitted to a method that tracks it.
These differences are not erased by reinforcement learning, since policies reproduce the bias of their reference. DDR policies are deployed zero-shot, with the same pipeline and weights, on two physical humanoids of very different sizes, the full-size H1-2 and the small R1, including a crawl and a side kick whose contact sequences are specified nowhere in the pipeline. We also release a dataset of DDR references covering all the flat-ground motions of AMASS on three humanoids.

When an accurate geometric reference is available, tracking its keypoints in task space is as accurate as DDR, so the benefit of tracking the human demonstration is larger for noisy references such as monocular reconstructions. DDR also has limitations. It uses a geometric solution to warm-start the solver, and the keypoint mapping, the scaling across morphologies and the horizon of the Monocular-Video motions are chosen by hand. Its computational cost is about 3\,min per second of motion on H1-2 with one H100 GPU, and it is evaluated on flat ground. These limitations suggest clear next steps: a systematic study of the warm start and of the solver parameters, an automatic mapping and scaling across morphologies, large collections of in-the-wild videos, interactions with terrain and objects, and a lower computational cost, e.g., by learning to amortize the optimization.
 \section*{Acknowledgment}
This project was provided with computing HPC and storage resources by GENCI at IDRIS thanks to the grant 2026-AD011016104R1 on the supercomputer Jean Zay's V100 and H100 partitions. 
\bibliographystyle{IEEEtran}

\end{document}